\documentclass{article}
\usepackage{ijcai24}

\usepackage{times}
\usepackage{soul}
\usepackage{url}
\usepackage[hidelinks]{hyperref}
\usepackage[utf8]{inputenc}
\usepackage[small]{caption}
\usepackage{graphicx}
\usepackage{amsmath}
\usepackage{amsthm}
\usepackage{booktabs}
\usepackage{algorithm}
\usepackage{algorithmic}
\usepackage[switch]{lineno}
\usepackage{graphicx}
\usepackage{xr-hyper}
\usepackage{hyperref}
\usepackage{url}
\usepackage{multirow}
\usepackage{colortbl}
\usepackage{natbib} 
\usepackage{bm}

\usepackage{newfloat}
\usepackage{listings}
\usepackage{amssymb}
\usepackage{booktabs}
\usepackage{tabularray}
\usepackage{icomma}
\usepackage{stfloats}

\makeatletter
\newcommand*{\addFileDependency}[1]{
\typeout{(#1)}
\@addtofilelist{#1}
\IfFileExists{#1}{}{\typeout{No file #1.}}
}\makeatother

\newcommand*{\myexternaldocument}[1]{%
\externaldocument{#1}%
\addFileDependency{#1.tex}%
\addFileDependency{#1.aux}%
}

\myexternaldocument{appendix}

\title{DISTILLING ADVERSARIAL ROBUSTNESS USING HETEROGENEOUS TEACHERS}

\author{
Jieren Deng
\and
Aaron Palmer\and
Rigel Mahmood\and
Ethan Rathbun\and
Jinbo Bi \and
\\
Kaleel Mahmood \and
Derek Aguiar
\affiliations
University of Connecticut
\emails
\{jieren.deng, aaron.palmer, rigel.mahmood, ethan.rathbun, jinbo.bi, kaleel.mahmood, derek.aguiar\}@uconn.edu,
}

\begin{document}

\maketitle

\begin{abstract}
Achieving resiliency against adversarial attacks is necessary prior to deploying neural network classifiers in domains where misclassification incurs substantial costs, e.g. self-driving cars or medical imaging.
Recent work has demonstrated that robustness can be transferred from an adversarially trained teacher to a student model using knowledge distillation.
However, current methods perform distillation using a single adversarial and vanilla teacher and consider homogeneous architectures (i.e., CNNs) that are susceptible to misclassify examples from similar adversarial subspaces.  
In this work, we develop a defense framework against adversarial attacks by distilling adversarial robustness using heterogeneous teachers (DARHT), where teachers are different with respect to architecture or adversarial training algorithm.
In DARHT, the student model explicitly represents teacher logits in a student-teacher feature map and leverages multiple teachers that exhibit low adversarial example transferability (i.e., robust to different adversarial subspaces).
Experiments on classification tasks in both white-box and black-box scenarios demonstrate that DARHT exhibits state-of-the-art performance; in particular, DARHT achieved at least 3.26\%, 3.66\%, and 25.75\% improvements in weighted robust accuracy for Auto-PGD and 5.23\%, 4.76\%, and 22.83\% for Square Attack when compared to competing adversarial distillation methods in the CIFAR-10, CIFAR-100, and Tiny ImageNet datasets, respectively.
Comparisons between homogeneous and heterogeneous teacher sets suggest that leveraging teachers with low adversarial example transferability increases student model robustness. 
\end{abstract}

\section{Introduction}

As deep learning is deployed for an increasing variety of applications, the threat posed by adversarial examples becomes ever more prevalent. 
Machine learning classifiers (in particular deep neural networks) are especially vulnerable~\citep{szegedy2013intriguing, barz}. 
The consequences for misclassification in response to adversarial perturbation can be substantial in applications where safety is paramount~\citep{dalvi2004adversarial}, including sensor attacks on self-driving cars~\citep{cao2019adversarial} and misclassification of disease states from medical imaging~\citep{finlayson2019adversarial}. 
Thus, achieving resiliency against adversarial attacks is essential before deploying machine learning in technologies that interface with humans.

In general, there are two ways in which adversarial attacks can be mitigated. 
First, provably robust \textit{certified defenses} defend against attacks assuming that adversarial perturbations are bounded by a specified norm~\citep{raghunathan2018certified}. 
Second, attacks can be mitigated through the use of \textit{empirical defenses} like adversarial training~\citep{BART, madry2018towards}. 
While strategies to circumvent some empirical defenses have demonstrated success~\citep{tramer2020adaptive}, adversarial training techniques have shown a remarkable robustness to a diverse set of attacks. 
In this work, we focus on adversarial training~\citep{zhang2020fat}, as it is widely considered to be one of the most effective empirical defenses~\citep{madry2018towards, rice2020overfitting}. 
Adversarial training posits a robust modified loss function, and then iteratively perturbs a randomly selected training sample to have large loss before updating model parameters during training~\citep{bai2021recent}. 

Recent work has demonstrated that knowledge distillation, which is typically used to train neural networks in resource constrained settings~\citep{hinton2015distilling}, can be used to transfer adversarial robustness to a student model.
\textit{Defensive distillation} uses the predicted probabilities of a sample from a teacher network as training labels for a distilled network to improve generalization performance~\citep{papernot2016distillation}. 
In contrast, \textit{adversarially robust distillation} minimizes a discrepancy between the teacher's predicted class of an unperturbed example and the student's prediction on the same example after adversarial perturbation~\citep{goldblum2020adversarially,zhu2022reliable,zi2021revisiting}.
Subsequent methods incorporated an adversarial and vanilla (i.e., not adversarially trained) teacher explicitly in the distillation loss function~\citep{zhao2022enhanced, zi2021revisiting}.
Explanations for how adversarial knowledge is transferred appeal to leveraging inter-class correlations, encouraging predicted class probabilities with higher entropy by gradient scaling, label smoothing, and altering logit distributions~\citep{furlanello2018born,tang2020understanding,phuong2019towards,seguin2021understanding}.

In parallel to the developments in the field of adversarial knowledge distillation, another adversarial phenomena has been well documented, which is denoted as \textit{adversarial transferability}~\citep{liu2016delving, mahmood2021robustness, xu2022securing}. 
In this context transferability refers to the fact that models with similar architectures (e.g. two convolutional neural networks -- CNNs) often misclassify the same adversarial example. 
Adversarial examples span contiguous regions, or adversarial subspaces~\citep{goodfellow2014explaining}, and high transferability between two models can be explained by the extent to which these subspaces intersect.
While transferability exists in adversarial defense and distillation scenarios~\citep{papernot2016distillation}, models with different architectures (e.g. a CNN and vision transformer) do not misclassify the same set of adversarial examples when the samples are generated using single model white-box attacks like FGSM~\citep{goodfellow2014explaining}, PGD~\citep{madry2018towards} or CW$_{\infty}$~\citep{carlini2017towards}.
This raises an important question: \textit{can teachers with diverse architectures provide better robustness in the knowledge distillation setting?}

\begin{figure*}[h!]
\centering
\includegraphics[width=1\textwidth]{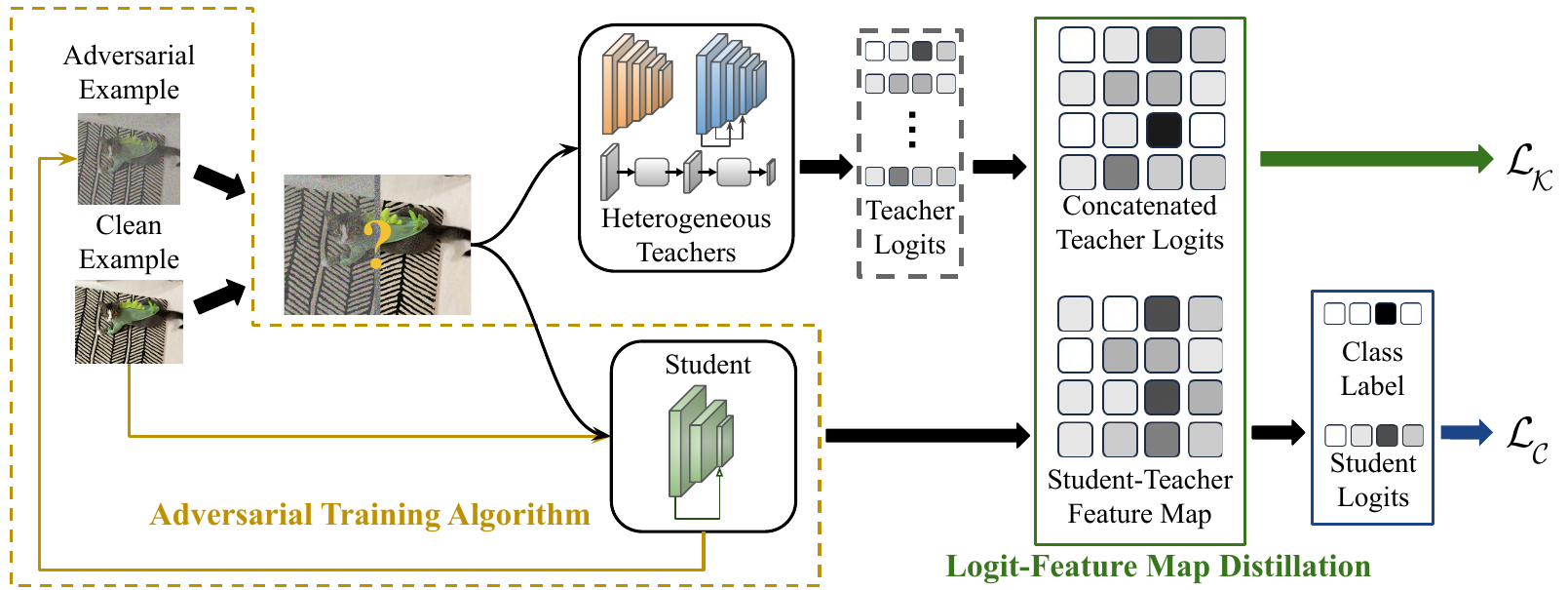}
\caption{\textbf{Distilling adversarial robustness using heterogeneous teachers (DARHT).} 
Adversarial examples are generated by applying an adversarial training algorithm to the clean examples using the student model.
Subsequently, DARHT selects adversarial or clean examples as input for both the student model and heterogeneous teachers. 
The knowledge distillation loss ($\mathcal{L}_\mathcal{K}$) is calculated using the teacher logits and the student-teacher feature map. 
The classification loss ($\mathcal{L}_\mathcal{C}$) is computed as the cross-entropy between the student logits and the labels.
}
\label{Fig:overall} 
\end{figure*}


In this work, we introduce a defense framework against white-box adversarial attacks by  distilling adversarial robustness using heterogeneous teachers (DARHT; Fig.~\ref{Fig:overall}).
Unlike prior work, DARHT leverages both adversarially trained and vanilla teacher models that exhibit low adversarial transferability.
The student model explicitly represents the teacher logits, whose properties have been shown to be valuable for adversarial robustness~\citep{seguin2021understanding}, in a \textit{student-teacher feature map}.
DARHT accommodates heterogeneous teachers by learning weights for each individual student-teacher feature and avoids expensive per-teacher hyperparameter tuning by weighing a specific teacher's distillation loss by the inverse of its cross-entropy loss.
In total, our work makes the following contributions:

\begin{itemize}
    \item We develop DARHT, the first approach for adversarial knowledge distillation with multiple heterogeneous teachers\footnote{The source code for \texttt{DARHT} is available at \href{https://github.com/bayesomicslab/DARHT}{https://github.com/bayesomicslab/DARHT}.}.
    \item We show that using teacher models that exhibit low adversarial transferability (or low adversarial subspace intersection~\citep{tramer2017space})  increases adversarial robustness of the student in a knowledge distillation setting, an effect we term \textit{complementary robustness}.
    \item Extensive classification experiments in both white-box and black-box settings demonstrate that DARHT achieved at least 3.26\%, 3.66\%, and 25.75\% improvements in weighted robust accuracy (the average of robust and clean accuracies) for Auto-PGD~\citep{croce2020reliable} and 5.23\%, 4.76\%, and 22.83\% for Square Attack~\citep{andriushchenko2020square} when compared to two state-of-the-art adversarial distillation approaches (RSLAD~\citep{zi2021revisiting} and MTARD~\citep{zhao2022enhanced}) in the CIFAR-10, CIFAR-100, and Tiny ImageNet datasets, respectively.    
    \item Further experiments demonstrate that robust and clean accuracies improve with (a) more teachers, (b) architecturally diverse teachers, and (c) teachers trained by different adversarial training algorithms. 
\end{itemize}

\section{Prior Work on Adversarial Distillation}
Adversarial robustness can be distilled to a student model using an adversarially trained teacher~\citep{bai2020feature,zhu2022reliable,goldblum2020adversarially,chen2020robust}.
Adversarially robust distillation (ARD) showed that robust student models can be produced without adversarially training the student~\citep{goldblum2020adversarially}; namely, ARD reformulates the adversarial training minimization problem in~\cite{madry2018towards} to focus the student on reproducing the teacher outputs within a closed ball of radius $\epsilon$. 
The Robust Soft Label Adversarial Distillation (RSLAD) method demonstrated improvements on ARD by incorporating robust soft labels (i.e., the softmax output of an adversarially trained teacher model)~\citep{zi2021revisiting}.
While not producing the highest clean accuracy, RSLAD demonstrated higher robust accuracies when compared with ARD with respect to five attack algorithms. 
The Multi-Teacher Adversarial Robustness Distillation (MTARD) method distills knowledge from both an adversarially trained and a vanilla  teacher~\citep{zhao2022enhanced,zhao2023mitigating}.
By including a vanilla teacher in distillation, MTARD balances clean and robust accuracy, resulting in an improved weighted robust accuracy (W-Robust). 
However, while MTARD achieves high clean accuracy, it is seldom the most robust. 
MTARD also includes separate weight hyperparameters for the student and teacher models to balance robust and clean accuracies that require hyperparameter tuning. 

\section{DARHT Framework}

In DARHT, the student explicitly \emph{represents} teacher logits in a \textit{student-teacher feature map} and combines a classification loss (Fig.~\ref{Fig:overall}, blue arrow) with a distillation loss (Fig.~\ref{Fig:overall}, green arrow) between the softmaxed student-teacher feature map and the softmaxed teacher logits for transferring adversarial robustness.
In this section, we describe necessary notations and the core components of DARHT: logit-feature map knowledge distillation, teacher heterogeneity, Monte Carlo dropout, and the training algorithm. 


\paragraph{Notation.}
We will consider classification problems with $K$ classes. The student model, $S_{\theta}$, is parameterized by $\theta$ and the set of teacher models is denoted by $T = \{T_j\}_{j=1}^J$. Applying the student model and teacher model $T_j$ to an example $x$ produces logit vectors $S_{\theta}(x)$ and $T_j(x)$ taking values in $\mathbb{R}^K$. \textit{Logit-feature map distillation} is a function of the teacher logits and the student-teacher feature map, which is the student's \emph{representation} of the teacher's logits. The full student-teacher feature map is $S_{\theta}^{(T)}(x) = \{S_{\theta}^{(T_{j})}(x)\}_{j=1}^J$, where $S_{\theta}^{(T_{j})}(x)$ represents the internal student-teacher feature map for the $j^{th}$ teacher for all $K$ classes.

\begin{figure}[t]
    \centering
    \includegraphics[width=0.8\columnwidth]{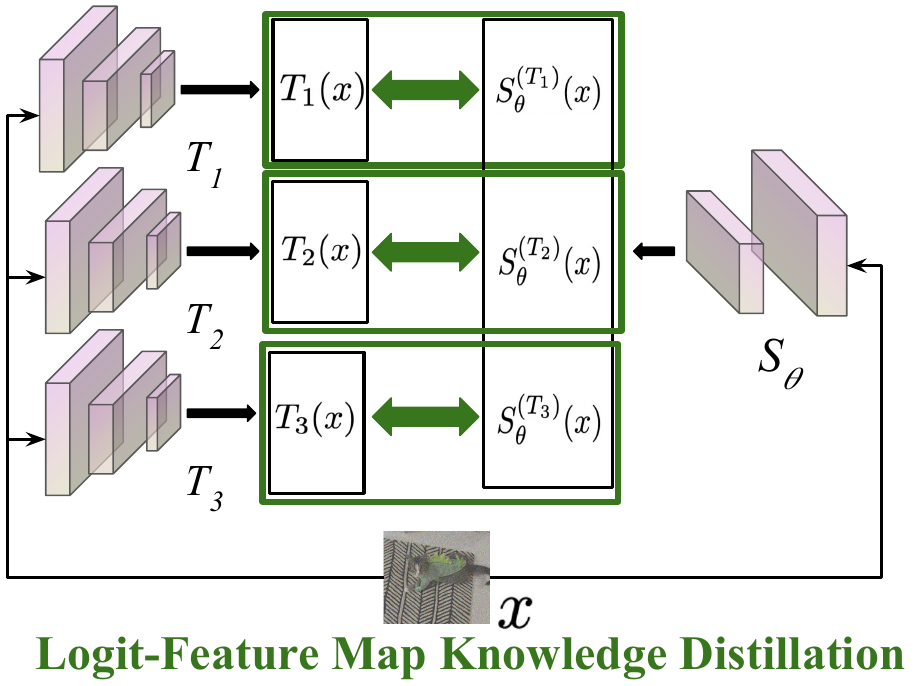}
    \caption{\textbf{Distilling knowledge from teacher logits to student-teacher feature map.}
    An example ($x$) passed to the student ($S_{\theta}$) and three teachers ($T_{1}$, $T_{2}$, and $T_{3}$) models.}  
    \label{fig:cross-feature}
\end{figure}

Training the student model involves an outer minimization of the population risk~\citep{madry2018towards} where an adversarial example $x_{adv}$ (with label $y$) is generated by adversarial training algorithm $\mathcal{F}$ using a clean example $x_{cln}$ based on the student's current $\theta$, i.e. $x_{adv} = \mathcal{F}(\theta,x_{cln},y)$. A forward pass through the student model produces the tuple $(S_{\theta}(x), S_{\theta}^{(T)}(x))$. The student objective is to minimize a combination of the classification distillation losses:
\begin{equation}
\operatorname*{arg\,min}_{\theta} \left(\mathcal{L}_{\mathcal{C}}\left( S_\theta, x, y \right) + \mathcal{L}_{\mathcal{K}}\left(S_{\theta}^{(T)}, T ,x \right)\right)
\label{ourobj}
\end{equation}
where $\mathcal{L}_{\mathcal{K}}$ and $\mathcal{L}_{\mathcal{C}}$ respectively indicate the knowledge distillation and classification loses. The classification loss, $\mathcal{L}_{\mathcal{C}}$, is defined as the cross-entropy between the softmaxed output of example $x$ and one-hot encoded label $y$:
\begin{equation*}
\mathcal{L}_{\mathcal{C}}\left( S_\theta, x, y \right) = -\sum_{k=1}^K y_{k}\log\left( \sigma_k(S_{\theta}(x))\right)
\end{equation*}
where $\sigma(\cdot)$ denotes the softmax function and $\sigma_k(\cdot)$ denotes the $k^{th}$ element. The distillation loss, illustrated in Figure~\ref{fig:cross-feature}, is given by a weighted combination of Kullback-Leibler divergences between the $j^{th}$ teacher's softmaxed output and corresponding student-teacher's softmaxed feature map:
\begin{equation*}
\label{eqn:kd}
\mathcal{L}_{\mathcal{K}}\left( S_{\theta}^{(T)},T,x \right) = \sum_{j=1}^J  w_j \cdot \tilde{\mathcal{L}}_{\mathcal{K}}\left( S_{\theta}^{(T_{j})},T_{j},x \right)
\end{equation*}
where 
\begin{equation*}
\label{eqn:kl}
\tilde{\mathcal{L}}_{\mathcal{K}}\left( S_{\theta}^{(T_{j})},T_{j},x \right) = \sum_{k=1}^{K}\sigma_k(S_{\theta}^{(T_{j})}(x)) \log{\frac{\sigma_k(S_{\theta}^{(T_{j})}(x))}{\sigma_k(T_{j}(x))}}.
\end{equation*}
The normalization weights allow teachers to contribute to the distillation loss inversely proportional to their cross-entropy loss and are defined by:
\begin{equation*}
w_j=\frac{\exp  \left( -\mathcal{L}_{\mathcal{C}}\left( T_{j}, x, y\right) \right) }{\sum_{j=1}^J \exp  \left( -\mathcal{L}_{\mathcal{C}}\left( T_{j}, x, y\right) \right) } .
\end{equation*}

\subsection{Teacher Heterogeneity}

Adversarial examples that are generated with respect to one model are often misclassified by another model with similar architectures~\citep{mahmood2021robustness}. This phenomena is referred to as transferability~\citep{liu2016delving}. However, adversarial examples generated for a specific model architecture  are not often misclassified by models with markedly different architectures~\citep{mahmood2021robustness}. For example, the adversarial examples generated by vision transformers (ViT)~\citep{dosovitskiy2020vit} are not often misclassified by ResNets. In the context of adversarial distillation, achieving robustness has focused primarily on CNNs~\citep{zhao2022enhanced,zi2021revisiting,zhu2022reliable,goldblum2020adversarially}. To the best of our knowledge, it remains an open question on how the low transferability phenomena occurring between CNNs and ViTs can be leveraged to achieve robustness in a knowledge distillation framework. DARHT specifically addresses this question by employing heterogeneous teachers (i.e., CNN and ViTs) to achieve greater robustness than a homogeneous teacher setup.  We include ViTs in the context of distilling adversarial robustness from multiple teachers, specifically to investigate whether multiple teachers with heterogeneous architectures provide complementary robustness to the student; that is, do the different learned features and decision boundaries of two significantly different architectures provide increased robustness compared to their homogeneous counterparts in a distillation setting.
Given that adversarial training algorithms do perturb adversarial subspaces~\citep{tramer2017space}, DARHT also considers heterogeneity with respect to adversarial training algorithms.

\subsection{Monte Carlo Dropout in DARHT}
The student model also implements Monte Carlo dropout, which is typically applied during neural network training to increase generalization performance~\citep{srivastava2014dropout} or to model uncertainty~\citep{gal2016dropout}. 
Defense dropout algorithms use dropout during training and testing and have been shown to increase model robustness against adversarial attacks in a way that is complementary to distillation as defense~\citep{wang2018defensive}.
We implement Monte Carlo dropout on the feature map before the student-teacher feature map. 
By conducting several forward passes on a given example, the model generates a variety of predictions, which are subsequently averaged before the softmax operation.

\subsection{DARHT Training}

DARHT optimizes the Equation~\ref{ourobj} with respect to student parameters $\theta$ using Algorithm~\ref{alg:cap}.
Mini-batches are created by repeatedly sampling a clean example and generating an adversarial example with probability $0.5$ using an adversarial training algorithm and the student model.

\begin{algorithm}[H]
\caption{DARHT Training}\label{alg:cap}
\textbf{Input}: clean examples $x_{cln}$ and labels $y$\\
\textbf{Parameters}: Student model $S_{\theta}$, pretrained teacher models $\{T_{j}\}_{j=1}^J$, adversarial algorithm $\mathcal{F}(\cdot)$, 
training iterations $I$, Monte Carlo (MC) iterations $C$\\
\textbf{Output}: adversarially trained $S_{\theta}$
\begin{algorithmic}[1] 
  \FOR{epochs $i=1$ to $I$}
    \STATE $x_{adv}, y \gets \mathcal{F}(\theta_{s}^{(i)},x,y)$ \COMMENT{comp. adv. example}
    \STATE $x \gets \{ x_{adv}, x_{cln}\}$ \COMMENT{randomly pick example}
    \STATE{Initialize empty arrays $U, V$}
    \FOR{MC iterations $c=1$ to $C$}
        \STATE{Store $S_{\theta^{(i)}}(x)$ in $U$, $S_{\theta^{(i)}}^{(T)}(x)$ in $V$}
    \ENDFOR
    \STATE{$\bar{S}_{\theta^{(i)}}(x), \bar{S}_{\theta^{(i)}}^{(T)}(x) \gets \frac{1}{C}\sum_{c=1}^C U_c, \frac{1}{C}\sum_{c=1}^C V_c$} 
    \STATE $\mathcal{L} \gets \mathcal{L}_{\mathcal{C}}\left( \bar{S}_{\theta^{(i)}}, x, y \right) + \mathcal{L}_{\mathcal{K}}\left(\bar{S}_{\theta^{(i)}}^{(T)}, T ,x \right) $  
    \STATE $\theta_{i+1} \gets \theta_{i}-\eta\nabla_{\theta_i}\cdot \mathcal{L}$ 
\ENDFOR
\end{algorithmic}
\end{algorithm}

%

\section{Results}


\paragraph{Model and Datasets.} 
We performed extensive evaluations of DARHT with respect to both adversarial distillation methods and adversarial training algorithms on the CIFAR-10, CIFAR-100~\citep{krizhevsky2009learning} and Tiny ImageNet~\citep{Le2015TinyIV} datasets. 
For comparison we employed two state-of-the-art adversarial distillation approaches, RSLAD~\citep{zi2021revisiting} and MTARD~\citep{zhao2022enhanced}. RSLAD and MTARD are trained with a student ResNet-18~\citep{He2015DeepRL}  model.
We also compared DARHT with four different adversarial training algorithms, Madry~\citep{madry2018towards}, curriculum adversarial training (CAT)~\citep{cai2018curriculum}, dynamic adversarial training (DAT)~\citep{wang2019convergence}, friendly adversarial training (FAT)~\citep{zhang2020fat}. Madry, CAT, DAT and FAT are trained with a student WideResNet-32 and WideResNet-34~\citep{Zagoruyko2016WideRN}.

\paragraph{Training and Evaluation.} 
For the CIFAR-10 and CIFAR-100 datasets, we trained student models using a stochastic gradient descent (SGD) optimizer and 100 epochs with an initial learning rate, momentum, and weight decay set to $0.1$, $0.9$, and 2e-4, respectively. 
For Tiny ImageNet, we used a linear learning rate warmup to reach a learning rate $0.1$ at the $20^{th}$ epoch and then a learning rate decay scheduled at epochs $60$ and $80$. 
We evaluated the adversarial robustness of defense strategies with respect to clean examples and adversarial examples generated from white-box attacks, including the fast gradient sign method (FGSM)~\citep{goodfellow2014explaining}, projected gradient descent (PGD)~\citep{madry2018towards}, Carlini and Wagner using an $L_\infty$ distance metric (CW$_\infty$)~\citep{carlini2017towards}, and Auto-PGD~\citep{croce2020reliable}, and the black-box Square Attack~\citep{andriushchenko2020square}.
We used a $10$-step PGD with random start size $0.001$ and step size $2/255$. 
The maximum perturbation was bounded to the $L_{\infty}$ norm $\epsilon = 8/255$ for all attacks.

Adversarial examples used in training were generated following FAT~\citep{zhang2020fat} and the tradeoff-inspired adversarial dEfense via surrogate-loss minimization (TRADES) method~\citep{zhang2019theoretically} original settings; in direct comparisons with other adversarial distillation approaches, we used the same adversarial models.
We evaluated adversarial distillation approaches using the weighted robust accuracy (W-Robust), which is the average of the robust and clean accuracies~\citep{zhao2022enhanced}.
Our appendix provides detailed information about the heterogeneous teacher models we used, including ResNet-164s, vision transformers (ViTs), and WideResNets (WRNs).
\begin{figure*}[t!]
\centering
    \includegraphics[width=2\columnwidth]{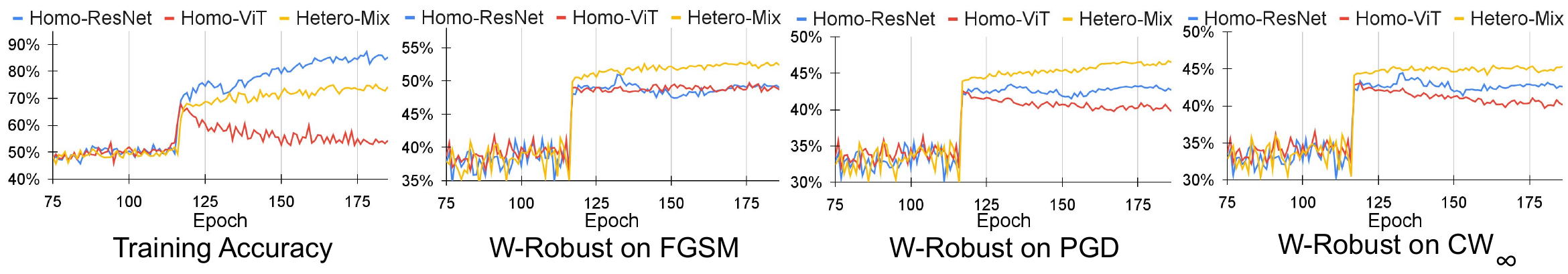}
    \caption{\textbf{The performance of the student (ResNet-18) on training accuracy and W-Robust metrics under FGSM, PGD, and CW$_\infty$ attacks is evaluated with three distinct groups of teachers: Homo-ResNet, Hetero-Mix, and Homo-ViT on the CIFAR-100 dataset.}}
    \label{fig:three_teachers_exp}
\end{figure*}

\subsection{Evaluating Heterogeneous Teachers}

First, we evaluated whether training with architecturally diverse teachers (low transferability configuration) provides some benefit with respect to robustness compared to architecturally similar teachers (high transferability).
We created three DARHT configurations that vary based on their teacher architecture configuration and performance on CIFAR-100 (Table~\ref{tab:three_hete_teachers}).
We tested the training accuracy and weighted robustness on CIFAR-100 for each DARHT configuration (Figure~\ref{fig:three_teachers_exp}).
The Hetero-Mix DARHT (heterogeneous teachers) had consistently better weighted robustness than the homogeneous DARHT configurations for each attack.
Interestingly, although the Homo-ResNet configuration had the highest training accuracy, its weighted robustness was only marginally better than the worst performing configuration that only used natural teachers (Homo-ViT).
This may be due to the high adversarial transferability between the teacher (ResNet-164) and student architectures (ResNet-18)~\citep{yu2023reliable}, or the exceedingly high accuracy of the teachers leading to overfitting in the student, which is suggested by the increasing training accuracy with simultaneously stable weighted robustness. 
Overall, these results suggest that heterogeneous teacher architectures do provide increased robustness compared to homogeneous teachers (i.e., provide complementary robustness).

\begin{figure}[h!tbp]
    \centering
    \includegraphics[width=1\columnwidth]{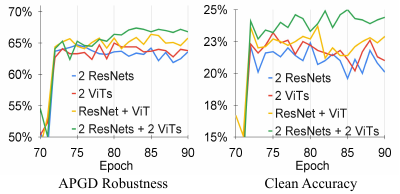}
    \caption{\textbf{APGD Robustness and clean accuracies for the DARHT student (ResNet-18) and different teacher compositions across epochs on CIFAR-100.}}
    \label{fig:hete_homo_teachers}
\end{figure}

\begin{table}[h!]
\centering
\begin{tblr}{
  cells = {c},
  cell{2}{1} = {r=2}{},
  cell{4}{1} = {r=2}{},
  cell{6}{1} = {r=2}{},
  hline{1,8} = {-}{0.08em},
  hline{2,4,6} = {-}{},
}
Configurations & Models        & Accuracy\\
Homo-ResNet      & AT-ResNet-164$_1$ & \textbf{99.79\%}                  \\
              & AT-ResNet-164$_2$ & 99.47\%                           \\
Hetero-Mix     & AT-WRN-70     & 88.56\%                           \\
              & AT-ViT-B-16   & 75.56\%                           \\
Homo-ViT    & NAT-ViT-B-16  & 60.10\%                           \\
              & NAT-ViT-B-32  & 71.92\%                           
\end{tblr}
\caption{\textbf{Teacher's details for Homo-ResNet, Hetero-Mix and Homo-ViT groups.} This categorization of the groups was done based on the accuracy and architecture of the teachers on the student training data (CIFAR-100).}

\label{tab:three_hete_teachers}
\end{table}
\paragraph{Additional teachers.}
Next, we evaluated the clean and robust accuracies of DARHT with respect to the FGSM, Auto-PGD (APGD), and CW$_{\infty}$ attacks as a function of the cardinality and composition of the teacher set on CIFAR-100.  
We included four DARHT configurations with different adversarially trained teacher compositions: (1) two WRN-70, (2) two ViT-B-32, (3) one WRN-70 and one ViT-B-32, and (4) two WRN-70 and two ViT-B-32. 
Teacher models were trained using FAT with $\tau=1$; for compositions that had two of the same model architecture, the second model was adversarially trained with $\tau=2$.
Based on our prior observations of under- and overfitting (Figure~\ref{fig:three_teachers_exp}), we selected teachers with accuracies between 75\% and 90\% on the training data. 
We again observed that DARHT configurations with heterogeneous teachers outperformed configurations with homogeneous teachers (Table~\ref{tab:hete_teachers}).
Moreover, the configuration with four teachers outperformed other DARHT configurations, exhibiting improvements of up to 2.5\% in clean accuracy and up to 6.1\% in adversarial robustness to APGD when compared to the second highest performing DARHT configuration. 
Additionally, the accuracies for clean examples and adversarial attacks were relatively stable over epochs (Figure~\ref{fig:hete_homo_teachers}).

\begin{table}[h!]
\centering
\resizebox{\columnwidth}{!}{
\centering
\begin{tblr}{
  column{2} = {c},
  column{3} = {c},
  column{5} = {c},
  column{6} = {c},
  vline{2-6} = {-}{},
  hline{1,6} = {-}{0.08em},
  hline{2} = {-}{},
}
Configurations  & Clean  & FGSM   & PGD    & APGD   & CW$_{\infty}$     \\
2 WRN           & 63.6\% & 32.9\% & 21.8\% & 20.1\% & 22.4\% \\
2 ViT         & 63.8\% & 36.1\% & 22.6\% & 21.0\% & 23.0\% \\
WRN + ViT     & 65.4\% & 35.1\% & 23.7\% & 22.9\% & 23.3\% \\
2 WRN + 2 ViT & \textbf{67.1\%} & \textbf{38.6\%} & \textbf{25.9\%} & \textbf{24.3\%} & \textbf{24.5\%} 
\end{tblr}
}
\caption{\textbf{DARHT accuracy with a ResNet-18 student on CIFAR-100 as a function of teacher composition.}
WRN and ViT denote WRN-70 and ViT-B-32 architectures, respectively.}
\label{tab:hete_teachers}.
\end{table}

\paragraph{Transferability experiments.}
We next evaluated whether heterogeneous teachers provide an explicit advantage by reducing transferability of adversarial examples from the student to the teachers.
In total, we evaluated three DARHT configurations: (1) two ResNet-164, (2) one WRN-70 and one ViT-B-16, and (3) two ViT-B-16.
We randomly selected 1000 examples from CIFAR-100 that all teachers classified correctly and generated adversarial examples using their respective DARHT student models (ResNet-18).
Then, we evaluated transferability by computing the number of adversarial examples that fooled the student and the teacher normalized by the number of examples fooling the student (i.e., examples that transferred to the teacher).
Transferability for the homogeneous DARHT configurations (1) and (3) were $46.48\%$ and $49.81\%$, respectively; only $39.58\%$ of adversarial examples transferred in the heterogeneous DARHT configuration.
In the same heterogeneous model, we also noticed that $27.05\%$ of examples that the ViT-B-16 model misclassified were correctly classified by the student; this is considerably higher than the next highest model ($7.91\%$ in configuration 3).
One possible explanation for this is that training with the WRN-70 model helps the student recover from the (erroneous) learned knowledge of the ViT-B-16 model by exploiting the low transferability between these architectures. 
In total, these experiments suggest that leveraging a more diverse set of teachers decreases adversarial transferability from students to teachers and, combined with a student that explicitly represents teacher logits, increases adversarial robustness.

\subsection{Comparing Adversarial Distillation Methods}
\label{sec:distill}

Having established that DARHT with heterogeneous teachers is preferred, we evaluated a heterogeneous DARHT configuration against two state-of-the-art adversarial distillation approaches, MTARD and RSLAD, on CIFAR-100.
Since MTARD uses two teacher models, we selected a two-teacher configuration for DARHT, which included the same WRN-70 adversarial teacher used in MTARD and RSLAD (trained with TRADES) and a ViT-B-16 (trained with FAT).
The student architecture was a ResNet-18 for all methods.
We evaluated performance based on clean accuracy, robust accuracy and W-Robust with respect to white-box attacks FGSM, PGD, CW$_{\infty}$, and APGD, and the black-box Square Attack.

\begin{figure}[h!tbp]
    \centering
    \includegraphics[width=0.9\columnwidth]{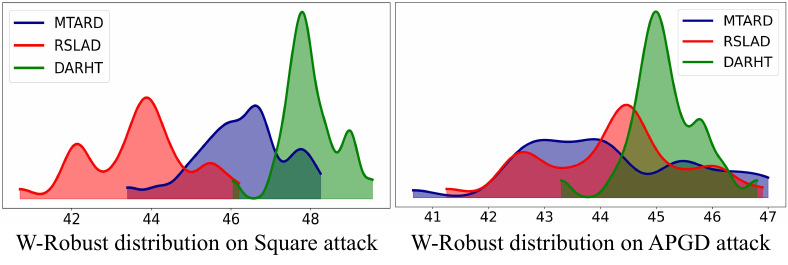}
    \caption{\textbf{W-Robust distributions with respect to the APGD and Square attacks for MTARD, RSLAD and DARHT.}}
    \label{fig:w_robust_distribution}
\end{figure}

DARHT achieved the highest clean and W-Robust accuracies across all adversarial attack methods.
In particular, DARHT performed 9.6\% and 8.0\% better than RSLAD and MTARD, respectively, in terms of W-Robust and with respect to the CW$_{\infty}$ attack (Table~\ref{cifar_10_100}).
While RSLAD achieved the best robust accuracies across most adversarial attacks, the clean accuracy was substantially lower than MTARD and DARHT.
With respect to the black-box Square Attack, DARHT achieved 10.8\% and 4.76\% improvement in W-Robust compared to RSLAD and MTARD, respectively.
Next, we executed the APGD and Square attacks on each model using 30 trials each with 1,000 adversarially generated examples. 
DARHT achieved significantly higher W-Robust than RSLAD for both SA and APGD (Mann–Whitney $U$ test, $p \leq  1.66\times 10^{-11}$ and $p \leq 2.66 \times 10^{-3}$, respectively); similar results were observed with respect to MTARD (Mann–Whitney $U$ test, $p \leq 5.08 \times10^{-8}$ and $p \leq 2.70 \times 10^{-4}$, respectively).
DARHT's increased performance was evidenced by the W-Robust distributions of DARHT being shifted to the right when compared to RSLAD and MTARD (Figure~\ref{fig:w_robust_distribution}). 
Results generated for CIFAR-10 were largely consistent with these evaluations (see Appendix).

\begin{table}[h!tbp]
\centering
\setlength{\extrarowheight}{0pt}
\addtolength{\extrarowheight}{\aboverulesep}
\addtolength{\extrarowheight}{\belowrulesep}
\setlength{\aboverulesep}{0pt}
\setlength{\belowrulesep}{0pt}
\begin{tabular}{llrrr} 
\toprule
\multirow{2}{*}{Attack}                   & \multirow{2}{*}{Defense} & \multicolumn{3}{c}{CIFAR-100}                                                                              \\ 
\cline{3-5}
                                          &                          & \multicolumn{1}{l}{Clean} & \multicolumn{1}{l}{Robust} & \multicolumn{1}{l}{W-Robust}                      \\ 
\hline
\multicolumn{1}{c}{\multirow{3}{*}{FGSM}} & RSLAD                    & 58.25\%                   & 34.73\%                    & {\cellcolor[rgb]{0.988,0.988,1}}46.49\%           \\
\multicolumn{1}{c}{}                      & MTARD                    & 64.30\%                   & 31.49\%                    & {\cellcolor[rgb]{0.988,0.988,1}}47.90\%           \\
\multicolumn{1}{c}{}                      & DARHT                    & \textbf{66.98\%}          & \textbf{40.10\%}           & {\cellcolor[rgb]{0.988,0.988,1}}\textbf{53.54\%}  \\ 
\hline
\multirow{3}{*}{PGD}                      & RSLAD                    & 58.25\%                   & \textbf{31.19\%}           & {\cellcolor[rgb]{0.988,0.988,1}}44.72\%           \\
                                          & MTARD                    & 64.30\%                   & 24.95\%                    & {\cellcolor[rgb]{0.988,0.988,1}}44.63\%           \\
                                          & DARHT                    & \textbf{66.98\%}          & 27.90\%                    & {\cellcolor[rgb]{0.988,0.988,1}}\textbf{47.44\%}  \\ 
\hline
\multirow{3}{*}{CW$_{\infty}$}            & RSLAD                    & 58.25\%                   & \textbf{28.21\%}           & {\cellcolor[rgb]{0.988,0.988,1}}43.23\%           \\
                                          & MTARD                    & 64.30\%                   & 23.42\%                    & {\cellcolor[rgb]{0.988,0.988,1}}43.86\%           \\
                                          & DARHT                    & \textbf{66.98\%}          & 27.80\%                    & {\cellcolor[rgb]{0.988,0.988,1}}\textbf{47.39\%}  \\ 
\hline
\multirow{3}{*}{APGD$^{*}$}               & RSLAD                    & 57.91\%                   & \textbf{30.52\%}           & {\cellcolor[rgb]{0.988,0.988,1}}44.22\%           \\
                                          & MTARD                    & 64.27\%                   & 24.06\%                    & {\cellcolor[rgb]{0.988,0.988,1}}44.17\%           \\
                                          & DARHT                    & \textbf{66.98\%}          & 24.70\%                    & {\cellcolor[rgb]{0.988,0.988,1}}\textbf{45.84\%}  \\ 
\hline
\multirow{3}{*}{SA$^{*}$}                 & RSLAD                    & 57.91\%                   & 29.48\%                    & {\cellcolor[rgb]{0.988,0.988,1}}43.70\%           \\
                                          & MTARD                    & 64.27\%                   & 28.25\%                    & {\cellcolor[rgb]{0.988,0.988,1}}46.26\%           \\
                                          & DARHT                    & \textbf{66.98\%}          & \textbf{29.94\%}           & {\cellcolor[rgb]{0.988,0.988,1}}\textbf{48.46\%}  \\
\bottomrule
\end{tabular}
\caption{\textbf{Adversarial distillation method performance on CIFAR-100}. 
The results of MTARD and RSLAD are reported from ~\citep{zhao2022enhanced} and ~\citep{zi2021revisiting} for FGSM, PGD, and CW$_{\infty}$ attacks.
The * symbol denotes that the results for APGD and SA are reproduced using officially released models.}
\label{cifar_10_100}
\end{table}

\begin{table}[h!tbp]
\centering
\begin{tblr}{
  row{odd} = {c},
  row{4} = {c},
  row{6} = {c},
  row{8} = {c},
  row{10} = {c},
  row{12} = {c},
  row{14} = {c},
  cell{1}{1} = {r=2}{},
  cell{1}{2} = {r=2}{},
  cell{1}{3} = {c=3}{},
  cell{3}{1} = {r=4}{},
  cell{7}{1} = {r=4}{},
  cell{11}{1} = {r=4}{},
  hline{1,15} = {-}{0.08em},
  hline{2} = {3-5}{},
  hline{3,7,11} = {-}{},
}
Attack      & Defense & Tiny ImageNet    &                  &                  \\
            &         & Clean            & Robust           & W-Robust         \\
CW$_\infty$ & RSLAD   & 36.80\%          & 14.80\%          & 25.80\%          \\
            & MTARD   & 40.90\%          & 12.24\%          & 26.57\%          \\
            & LAS-AT  & 44.86\%          & \textbf{18.45\%} & 31.66\%          \\
            & DARHT   & \textbf{48.20\%} & 17.32\%          & \textbf{32.76\%} \\
APGD        & RSLAD   & 36.80\%          & 16.70\%          & 26.75\%          \\
            & MTARD   & 40.90\%          & 13.70\%          & 27.30\%          \\
            & LAS-AT  & 44.86\%          & \textbf{21.96\%} & 33.41\%          \\
            & DARHT   & \textbf{48.20\%} & 20.45\%          & \textbf{34.33\%} \\
SA          & RSLAD   & 36.80\%          & 15.60\%          & 26.20\%          \\
            & MTARD   & 40.90\%          & 13.60\%          & 27.25\%          \\
            & LAS-AT  & 44.86\%          & \textbf{19.80\%} & 32.33\%          \\
            & DARHT   & \textbf{48.20\%} & 18.74\%          & \textbf{33.47\%} 
\end{tblr}
\caption{\textbf{Clean and robust accuracies for adversarial distillation methods and LAS-AT under CW$_\infty$, APGD, and Square attacks on the Tiny ImageNet dataset.}}
\label{tab:tiny_imgnet}
\end{table}

We also assessed adversarial distillation methods on Tiny ImageNet.
Given the lack of an official implementation of MTARD and RSLAD, we reproduced MTARD and RSLAD using the same hyperparameter settings employed in DARHT; further implementation considerations are discussed in the appendix.
DARHT consistently outperformed RSLAD and MTARD in terms of clean accuracy, robustness, and W-Robust across all metrics (Table~\ref{tab:tiny_imgnet}). 
Even when compared to the state-of-the-art \textit{learnable attack strategy} adversarial training method (LAS-AT)~\citep{las-at}, DARHT achieved a higher clean accuracy and 2.7\%  and 3.5\% gains in W-Robust for the APGD and Square attacks, respectively. 

\subsection{Comparing Adversarial Training Algorithms}
\label{sec:advTraining}

We evaluated the efficacy of our adversarial distillation approach against four adversarial training algorithms on CIFAR-10 (Table.~\ref{Tab:comp_fat}).
We kept the same settings as prior work for both models, WRN-32 and WRN-34~\citep{zhang2020fat}; DARHT included a ViT-B-16 and WRN-34 teacher (both FAT trained).
Since adversarially distillation techniques can benefit from early stopping~\citep{maroto2022benefits}, we terminated adversarial training after $65$ epochs, which is approximately half the number of epochs ($120$) typically used for FAT.
Compared to other adversarial training methods, DARHT-WRN-32 achieved the best clean and robust accuracies against all attacks. 
With respect to FAT, which was used to adversarially train DARHT teachers, DARHT-WRN-32 achieved notable improvements of 9.69\% for PGD and 10.42\% for CW$_\infty$. 
Similar results were observed when comparing DARHT with FAT in WRN-34 models. 
These results highlight how (a) multiple teachers in an adversarial distillation framework can improve clean and robust accuracies compared to single-model adversarial training methods. 

\begin{table}[h!tbp]
\centering
\resizebox{0.45\textwidth}{!}{
\begin{tblr}{
  cells = {c},
  cell{1}{1} = {r=2}{},
  cell{1}{2} = {c=4}{},
  hline{1,10} = {-}{0.08em},
  hline{2-3} = {2-5}{},
  hline{8} = {-}{},
}
Defense              & Attack (\%) &               &               &               \\
                     & Clean       & FGSM          & PGD           & CW$_\infty$   \\
Madry-WRN-32$^*$     & 87.3        & 56.1          & 45.8          & 46.8          \\
CAT-WRN-32$^*$       & 77.4        & 57.1          & 46.1          & 42.3          \\
DAT-WRN-32$^*$       & 85.0        & 63.5          & 48.7          & 47.3          \\
FAT-WRN-32$^\dagger$ & 89.3        & 65.5          & 46.1          & 46.8          \\
DARHT-WRN-32         & \textbf{91.9}        & \textbf{70.8} & \textbf{50.6} & \textbf{51.7} \\
FAT-WRN-34$^\dagger$ & 89.9        & 61.0          & 49.7          & 49.4          \\
DARHT-WRN-34         & \textbf{91.7}        & \textbf{69.4} & \textbf{50.2} & \textbf{51.7} 
\end{tblr}}
\caption{\textbf{White-box attacks on DARHT and adversarially trained WRN for CIFAR-10.}  
The $*$ and $\dagger$ symbols denote results reported from \citet{pmlr-v97-wang19i} and \citet{zhang2020fat}, respectively.}
\label{Tab:comp_fat}
\end{table}



\subsection{Evaluating Heterogeneous Training Methods}

\begin{table}[h!tbp]
\centering
\resizebox{0.45\textwidth}{!}{
\begin{tblr}{
  cells = {c},
  hlines,
  hline{1,4} = {-}{0.08em},
}
Teacher Groups & Clean           & FGSM            & PGD             & CW$_\infty$     \\
Hetergeneous       & \textbf{91.9\%} & \textbf{70.8\%} & \textbf{50.6\%} & \textbf{51.7\%} \\
Homogeneous          & 91.6\%          & 68.8\%          & 49.9\%          & 50.3\%          
\end{tblr}}

\caption{\textbf{Accuracies of students in homogeneous and heterogeneous adversarial training methods for teachers on CIFAR-10.} The homogeneous teacher configuration includes a FAT trained ViT and a FAT trained WRN. The heterogeneous teacher configuration includes a TRADES trained WRN and a FAT trained ViT.}
\label{tab:hete_adv_table}
\end{table}

\begin{figure}[h!]
    \centering
    \includegraphics[width=0.9\columnwidth]{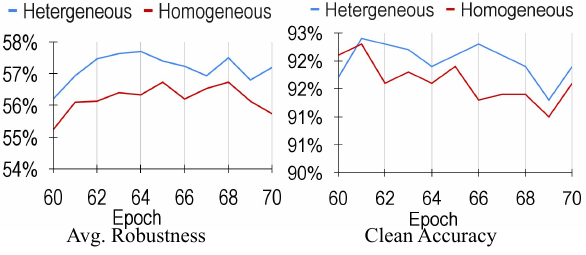}
    \caption{\textbf{DARHT performance across epochs for heterogeneous adversarial training methods.}
    Average robustness (left) and clean accuracy (right) were tracked across epochs 60 to 70 for DARHT configurations with teachers trained using the same (FAT, red) or different (TRADES and FAT, blue) adversarial training methods.}
    \label{fig:hete_adv}
\end{figure}

Finally, we evaluated the impact of teachers trained with different adversarial training algorithms (FAT and TRADES) on student robustness.
We compared DARHT configurations with WRN-34 and a ViT-B teachers; the homogeneous DARHT teachers were trained using FAT and the heterogeneous DARHT teachers were trained using both FAT and TRADES. 
The heterogeneous DARHT configuration exhibited marginally improved robust and clean accuracies compared to the homogeneous DARHT (Table~\ref{tab:hete_adv_table}). 
We observed that the heterogeneous model consistently maintained higher clean accuracy and robustness averaged over attacks (FGSM, PGD and CW$_{\infty}$) than the homogeneous model for epochs 61 through 70 (Fig.~\ref{fig:hete_adv}).
However, we note that the increase in performance is relatively small when compared to heterogeneous architectures.

\section{Conclusion}

In this work, we introduced DARHT, a framework for combining the complementary strengths of heterogeneous teachers with low adversarial example transferability into a single student model. 
Results on CIFAR-10, CIFAR-100, and Tiny ImageNet demonstrated improvements in adversarial robustness and clean accuracy with respect to recently developed adversarial training algorithms and both single and multiple teacher adversarial distillation methods.
We demonstrated that DARHT effectively leverages heterogeneous teacher (both with respect to architecture and adversarial training algorithms) to provide increased robustness compared to the homogeneous teacher setting (a phenomenon we labeled complementary robustness).  
In summary, using multiple teacher models that differ with respect to architecture (and, to a degree, adversarial training algorithms) is an effective strategy to increase both the robustness and clean accuracy of a student model in adversarial distillation settings. 


\bibliographystyle{named}
\bibliography{ijcai24}

\end{document}


\maketitle

\section{Teacher Models Configurations}
This section provides additional details about teachers used to evaluate the heterogeneous teacher configurations of DARHT (Section 4.1, \textit{Evaluating Heterogeneous Teachers}). 
We evaluated three teacher configurations of DARHT: Homo-ResNet, Hetero-Mix, and Homo-ViT (Table~\ref{tab:table_1_ext}). 
Both AT-ResNet-164$_{1}$ and AT-ResNet-164$_{2}$ architectures in the Homo-ResNet configuration are 164-layer ResNets,  adversarially trained with FAT~\citep{zhang2020fat} and each employing different hyperparameters, $\tau$. 
The AT-WRN-70 architecture is a 70-layer WideResNet adversarially trained with TRADES~\citep{zhang2019theoretically}; the model was accessed from Google DeepMind~\citep{deepmindRresearch}. 
Additionally, a vision transformer (AT-ViT-B-16), which was adversarially trained using FAT, is included in Hetero-Mix configuration with AT-WRN-70, along with two non-adversarially trained (or natural) vision transformers (NAT-ViT-B-16 and NAT-ViT-B-32) in the Homo-ViT configuration.


\begin{table}[h!]
\caption{\textbf{Supplementary information for teachers in Section 4.1, \textit{Evaluating Heterogeneous Teachers}.}
Teacher configurations were grouped based on their architectures and accuracies on the student training data (CIFAR-100). 
The hyperparameter $\tau$ is associated with FAT training, and instances marked as N/A indicate that the model does not use this hyperparameter. ``NAT" represents non-adversarial trained and ``AT" denotes adversarial trained models.}
\centering
\resizebox{0.85\columnwidth}{!}{%
\begin{tabular}{@{}lcc@{}}
\toprule
Configurations     & \multicolumn{2}{c}{Homo-ResNet}                     \\ \midrule
Models             & AT-ResNet-164$_{1}$               & AT-ResNet-164$_{2}$               \\
Training Method   & FAT                      & FAT                      \\
 $\tau$ & 1                        & 2                        \\
Accuracy           & 99.79\%                  & 99.47\%                  \\ \midrule
Configurations     & \multicolumn{2}{c}{Hetero-Mix}                      \\ \midrule
Models             & AT-WRN-70            & AT-ViT-B-16                 \\
Training Method   & TRADES                   & FAT                      \\
 $\tau$ & N/A                       & 1                        \\
Accuracy           & 88.56\%                  & 75.56\%                  \\ \midrule
Configurations     & \multicolumn{2}{c}{Homo-ViT}                        \\ \midrule
Models             & NAT-ViT-B-16                 & NAT-ViT-B-32                 \\
Training Method   & NAT & NAT \\
 $\tau$ & N/A                       & N/A                       \\
Accuracy           & 60.10\%                  & 71.92\%                  \\ \bottomrule
\end{tabular}}
\label{tab:table_1_ext}
\end{table}


\section{Implementation on Tiny ImageNet}
For the experiments with Tiny ImageNet, we aligned DARHT with the image processing configuration from the publicly available implementation of LAS-AT~\citep{las-at}.
Specifically, we used an SGD optimizer with a batch size of 256 and adopted a cosine warm-up learning rate strategy, starting  with an initial learning rate of 0.002 and gradually reaching a maximum learning rate of 0.1 over 20 epochs. 
Additionally, we incorporated a multistep learning rate decay at the 60${th}$ and 80${th}$ epoch, with a decay rate of 0.1.
Since RSLAD only uses a single teacher, we chose the ViT-L-16 model adversarially trained with FAT~\citep{zhang2020fat}. 
We used a the same 2-teacher configuration for both MTARD and DARHT: the aforementioned ViT-L-16 and a BigTransfer (BiT-M-R50) model adversarially trained with FAT~\citep{bigtransfer}.
Both ViT-L-16 and BiT-M-R50 used the same $\tau$ of 1.



\section{Student Details for Training and Inference}
For experiments on CIFAR-100 and CIFAR-10, we used a ResNet-18 student model with two teachers (Table~\ref{tab:trainable_ResNet-18}).
The model is composed of 51 layers and 11,291,532 trainable parameters. 
For student model configurations, the last layer (linear layer) is configured with an output matching the number of classes for the specific task, i.e., [1,10] for CIFAR-10 and [1,100] for CIFAR-100. 
The layer preceding the final layer is also a linear layer, configured with an output dimension matching the product of the number of classes and the number of teachers, i.e., [1,20] for 2 teachers on CIFAR-10 and [1,200] for 2 teachers on CIFAR-100.
Monte Carlo dropout is applied to the third last layer (before the penultimate linear layer) with a dropout rate of 0.25. 

\begin{table}[h!]
\centering
\resizebox{0.8\columnwidth}{!}{%
\begin{tabular}{rrr}
\hline
Layer (type)     & Output Shape     & Param \#   \\ \hline
Conv2d1         & [1, 64, 32, 32]   & 1,728      \\
BatchNorm2d-2    & [1, 64, 32, 32]  & 128        \\
Conv2d-3         & [1, 64, 32, 32]   & 36,864     \\
BatchNorm2d-4    & [1, 64, 32, 32]   & 128        \\
Conv2d-5         & [1, 64, 32, 32]   & 36,864     \\
BatchNorm2d-6    & [1, 64, 32, 32]   & 128        \\
BasicBlock-7     & [1, 64, 32, 32]   & 0          \\
Conv2d-8         & [1, 64, 32, 32]  & 36,864     \\
BatchNorm2d-9    & [1, 64, 32, 32]   & 128        \\
Conv2d10        & [1, 64, 32, 32]  & 36,864     \\
BatchNorm2d11   & [1, 64, 32, 32] & 128        \\
BasicBlock12    & [1, 64, 32, 32]   & 0          \\
Conv2d13        & [1, 128, 16, 16]  & 73,728     \\
BatchNorm2d14   & [1, 128, 16, 16]  & 256        \\
Conv2d15        & [1, 128, 16, 16]  & 147,456    \\
BatchNorm2d16   & [1, 128, 16, 16]  & 256        \\
Conv2d17        & [1, 128, 16, 16]  & 8,192      \\
BatchNorm2d18   & [1, 128, 16, 16]  & 256        \\
BasicBlock19    & [1, 128, 16, 16]  & 0          \\
Conv2d-20        & [1, 128, 16, 16]  & 147,456    \\
BatchNorm2d-21   & [1, 128, 16, 16]  & 256        \\
Conv2d-22        & [1, 128, 16, 16]  & 147,456    \\
BatchNorm2d-23   & [1, 128, 16, 16]  & 256        \\
BasicBlock-24    & [1, 128, 16, 16]  & 0          \\
Conv2d-25        & [1, 256, 8, 8]    & 294,912    \\
BatchNorm2d-26   & [1, 256, 8, 8]    & 512        \\
Conv2d-27        & [1, 256, 8, 8]    & 589,824    \\
BatchNorm2d-28   & [1, 256, 8, 8]    & 512        \\
Conv2d-29        & [1, 256, 8, 8]    & 32,768     \\
BatchNorm2d-30   & [1, 256, 8, 8]    & 512        \\
BasicBlock-31    & [1, 256, 8, 8]    & 0          \\
Conv2d-32        & [1, 256, 8, 8]    & 589,824    \\
BatchNorm2d-33   & [1, 256, 8, 8]    & 512        \\
Conv2d-34        & [1, 256, 8, 8]    & 589,824    \\
BatchNorm2d-35   & [1, 256, 8, 8]    & 512        \\
BasicBlock-36    & [1, 256, 8, 8]    & 0          \\
Conv2d-37        & [1, 512, 4, 4]    & 1,179,648  \\
BatchNorm2d-38   & [1, 512, 4, 4]    & 1,024      \\
Conv2d-39        & [1, 512, 4, 4]    & 2,359,296  \\
BatchNorm2d-40   & [1, 512, 4, 4]    & 1,024      \\
Conv2d-41        & [1, 512, 4, 4]    & 131,072    \\
BatchNorm2d-42   & [1, 512, 4, 4]    & 1,024      \\
BasicBlock-43    & [1, 512, 4, 4]    & 0          \\
Conv2d-44        & [1, 512, 4, 4]    & 2,359,296  \\
BatchNorm2d-45   & [1, 512, 4, 4]    & 1,024      \\
Conv2d-46        & [1, 512, 4, 4]    & 2,359,296  \\
BatchNorm2d-47   & [1, 512, 4, 4]    & 1,024      \\
BasicBlock-48    & [1, 512, 4, 4]    & 0          \\
Dropout-49       & [1, 512]          & 0          \\
Linear-50        & [1, 200]          & 102,600    \\
Linear-51        & [1, 100]          & 20,100     \\ \hline
\multicolumn{3}{l}{Total params: 11,291,532}     \\
\multicolumn{3}{l}{Trainable params: 11,291,532} \\
\multicolumn{3}{l}{Non-trainable params: 0}      \\ \hline
\end{tabular}}
\caption{Trainable parameters details for ResNet-18 student model on CIFAR-100, assuming two teachers.}
\label{tab:trainable_ResNet-18}
\end{table}

\section{Additional Results}

\subsection{Comparing Adversarial Distillation Methods on CIFAR-10}
In addition to CIFAR-100 and Tiny ImageNet, we compared DARHT with MTARD~\citep{zhao2022enhanced} and RSLAD~\citep{zi2021revisiting} on the CIFAR-10 dataset.
We used a ResNet-18 for the student model with a configuration that was consistent with the ResNet-18 on CIFAR-100 (Table~\ref{tab:trainable_ResNet-18}). 
Notably, Linear-50 and Linear-51 of the ResNet-18 on CIFAR-10 have parameter counts of 10,260 and 210, respectively, reflecting the fewer number of classes in CIFAR-10, with output shapes of [1, 20] and [1, 10]. 
DARHT demonstrated the best performance in both clean accuracy and W-Robust across four white-box attacks (FGSM, PGD, CW$_{\infty}$, and APGD) and the black-box Square Attack (Table~\ref{tab:comp_cifar_10}), consistent with the results on CIFAR-100 and Tiny ImageNet.
Specifically, DARHT demonstrated improvements of 5.9\% and 5.2\% over MTARD and RSLAD, respectively, in the case of the black-box square attack (SA) and exhibited a 4.0\% improvement over RSLAD and a 3.2\% improvement over MTARD in the context of the white-box attack APGD for W-Robust.

\begin{table}[h!]
\centering

\resizebox{0.9\columnwidth}{!}{%
\begin{tblr}{
  cell{1}{1} = {r=2}{},
  cell{1}{2} = {r=2}{},
  cell{1}{3} = {c=3}{c},
  cell{3}{1} = {r=3}{},
  cell{6}{1} = {r=3}{},
  cell{9}{1} = {r=3}{},
  cell{12}{1} = {r=3}{},
  cell{15}{1} = {r=3}{},
  hline{1,18} = {-}{0.08em},
  hline{2} = {3-5}{},
  hline{3,6,9,12,15} = {-}{},
}
Attack        & Defense & CIFAR-10         &                  &                  \\
              &         & Clean            & Robust           & W-Robust         \\
FGSM          & RSLAD   & 84.00\%          & 60.40\%          & 72.20\%          \\
              & MTARD   & 87.40\%          & 61.20\%          & 74.30\%          \\
              & DARHT   & \textbf{89.70\%} & \textbf{67.60\%} & \textbf{78.65\%} \\
PGD           & RSLAD   & 84.00\%          & \textbf{53.90\%} & 68.95\%          \\
              & MTARD   & 87.40\%          & 50.70\%          & 69.05\%          \\
              & DARHT   & \textbf{89.70\%} & 53.40\%          & \textbf{71.55\%} \\
CW$_{\infty}$ & RSLAD   & 84.00\%          & \textbf{52.70\%} & 68.35\%          \\
              & MTARD   & 87.40\%          & 48.60\%          & 68.00\%          \\
              & DARHT   & \textbf{89.70\%} & 49.90\%          & \textbf{69.80\%} \\
APGD$^{*}$    & RSLAD   & 84.37\%          & 52.50\%          & 68.44\%          \\
              & MTARD   & 87.10\%          & 50.80\%          & 68.95\%          \\
              & DARHT   & \textbf{89.70\%} & \textbf{52.70\%} & \textbf{71.20\%} \\
SA$^{*}$      & RSLAD   & 84.37\%          & 52.40\%          & 68.39\%          \\
              & MTARD   & 87.10\%          & 50.60\%          & 68.85\%          \\
              & DARHT   & \textbf{89.70\%} & \textbf{55.20\%} & \textbf{72.45\%}  \end{tblr}}
\caption{\textbf{Adversarial attacks on adversarial distillation methods on CIFAR-10}. A comparison between DARHT and other adversarial knowledge distillation methods using a ResNet-18 student on the CIFAR-10. The * symbol denote that the results are reproduced using officially released models.}
\label{tab:comp_cifar_10}
\end{table}

\subsection{Transferability Experiment Table}
Here, we provide additional data to support the transferability experiments.
The Homo-ResNet (two ResNet-164) and Hetero-Mix (one WRN-70 and one ViT-B-16) configurations are consistent with Table~\ref{tab:table_1_ext}.
The Homo-ViT-B-16 (two ViT-B-16) configuration consisted of two Vision Transformer (ViT-B-16) that are adversarially trained with FAT~\citep{zhang2020fat}, as detailed in Table~\ref{tab:homo-vit}.
We can obverse that the Hetero-Mix configuration has the lowest adversarial transferability; that is, Hetero-Mix had the fewest adversarial examples that fooled the student and the teacher normalized by the number of examples fooling the student (Table~\ref{tab:transfer}).

\begin{table}[ht!]
\centering
\begin{tabular}{@{}ccc@{}}
\toprule
\multicolumn{3}{c}{(1) Homo-ResNet (two ResNet-164)}                   \\ \midrule
ResNet164$_{1}$      & Teacher Correct & Teacher Wrong \\ \midrule
Student Correct & 269             & 7             \\
Student Wrong   & 392             & 332           \\ \midrule
ResNet164$_{2}$       & Teacher Correct & Teacher Wrong \\ \midrule
Student Correct & 261             & 15            \\
Student Wrong   & 383             & 341           \\ \midrule
\multicolumn{3}{c}{(2) Hetero-Mix (one WRN-70 and one ViT-B-16)}                    \\ \midrule
WRN-70          & Teacher Correct & Teacher Wrong \\ \midrule
Student Correct & 450             & 8             \\
Student Wrong   & 353             & 189           \\ \midrule
ViT-B-16        & Teacher Correct & Teacher Wrong \\ \midrule
Student Correct & 369             & 89            \\
Student Wrong   & 302             & 240           \\ \midrule
\multicolumn{3}{c}{(3) Homo-ViT-B-16 (two ViT-B-16)}                      \\ \midrule
ViT-B-16$_{1}$          & Teacher Correct & Teacher Wrong \\ \midrule
Student Correct & 173             & 31            \\
Student Wrong   & 364             & 432           \\ \midrule
ViT-B-16$_{2}$          & Teacher Correct & Teacher Wrong \\
Student Correct & 173             & 31            \\
Student Wrong   & 435             & 361           \\ \bottomrule
\end{tabular}
\caption{Correctness of teachers and students on 1,000 adversarial examples that were generated using the student models. The corresponding clean examples are class-balanced and selected from CIFAR-100 testset that are correctly classified by both teachers.}
\label{tab:transfer}
\end{table}

\begin{table}[ht!]
\centering
\resizebox{0.75\columnwidth}{!}{%
\begin{tabular}{@{}lcc@{}}
\toprule
Configurations     & \multicolumn{2}{c}{(3) Homo-ViT-B-16} \\ \midrule
Models             & ViT-B-16$_{1}$      & ViT-B-16$_{2}$      \\
Training Methods   & FAT           & FAT          \\
$\tau$ & 1             & 2            \\
Accuracy           & 75.56\%       & 82.42\%      \\ \bottomrule
\end{tabular}}
\caption{Teacher information for Homo-ViT-B-16 (two ViT-B-16). Both ViT-B-16 are adversarially trained with FAT~\citep{zhang2020fat} differing in the hyperparameter $\tau$. The two ViT-Bs achieved the accuracy of 75.56\% and 82.42\% on student training data (CIFAR-100).}
\label{tab:homo-vit}
\end{table}

\clearpage

\bibliographystyle{named}
\bibliography{ijcai24}